\documentclass[lettersize,journal]{IEEEtran}
\usepackage{amsmath,amsfonts}
\usepackage{algorithmic}
\usepackage{algorithm}
\usepackage{array}
\usepackage[caption=false,font=normalsize,labelfont=sf,textfont=sf]{subfig}
\usepackage{textcomp}
\usepackage{stfloats}
\usepackage{url}
\usepackage{verbatim}
\usepackage{graphicx}
\usepackage{multirow}
\usepackage{epstopdf}
\usepackage{amsthm}
\usepackage{color}
\newtheorem{myDef}{Definition}

\def\BibTeX{{\rm B\kern-.05em{\sc i\kern-.025em b}\kern-.08em
    T\kern-.1667em\lower.7ex\hbox{E}\kern-.125emX}}
\usepackage{balance}
\begin{document}
\title{Modeling Multi-Granularity Context Information Flow for Pavement Crack Detection}
\author{Junbiao Pang, Baocheng Xiong, Jiaqi Wu

\thanks{Manuscript created October, 2020; This work was developed by the IEEE Publication Technology Department. This work is distributed under the \LaTeX \ Project Public License (LPPL) ( http://www.latex-project.org/ ) version 1.3. A copy of the LPPL, version 1.3, is included in the base \LaTeX \ documentation of all distributions of \LaTeX \ released 2003/12/01 or later. The opinions expressed here are entirely that of the author. No warranty is expressed or implied. User assumes all risk.}}


\maketitle

\begin{abstract}
Crack detection has become an indispensable, interesting yet challenging task in the computer vision community. Specially, pavement cracks have a highly complex spatial structure, a low contrasting background and a weak spatial continuity, posing a significant challenge to an effective crack detection method. In this paper, we address these problems from a view that utilizes contexts of the cracks and propose an end-to-end deep learning method to model the context information flow. To precisely localize crack from an image, it is critical to effectively extract and aggregate multi-granularity context,  including the fine-grained local context around the cracks (in spatial-level) and the coarse-grained semantics (in segment-level). Concretely, in Convolutional Neural Network (CNN), low-level features extracted by the shallow layers represent the local information, while the deep layers extract the semantic features. Additionally, a second main insight in this work is that the semantic context should be an guidance to local context feature. By the above insights, the proposed method we first apply the dilated convolution as the backbone feature extractor to model local context, then we build a context guidance module to leverage semantic context to guide local feature extraction at multiple stages. To handle label alignment between stages, we apply the Multiple Instance Learning (MIL) strategy to align the high-level feature to the low-level ones in the stage-wise context flow. In addition, compared with these public crack datasets, to our best knowledge, we release the largest, most complex and most challenging Bitumen Pavement Crack (BPC) dataset. The experimental results on the three crack datasets demonstrate that the proposed method performs well and outperforms the current state-of-the-art methods.
\end{abstract}


\section{Introduction}

\IEEEPARstart{P}{avement} crack is one of the most potential yet important factors that endanger road safety, which constantly erode the internal structure of the road. Pavement cracks have produced a series of hazards, seriously affecting the life of the infrastructure, especially for the large number of highways. As a consequence, with the increasing length of roads, the accurate detection of cracks is increasingly becoming a vital method to not only maintain the safety of roads but also extend the life of roads. Efficient crack detection method is a practical requirement especially for bitumen pavement.

\begin{figure*}[t!]
\begin{center}
\subfloat[A sample]{\includegraphics[width = 0.333\textwidth]{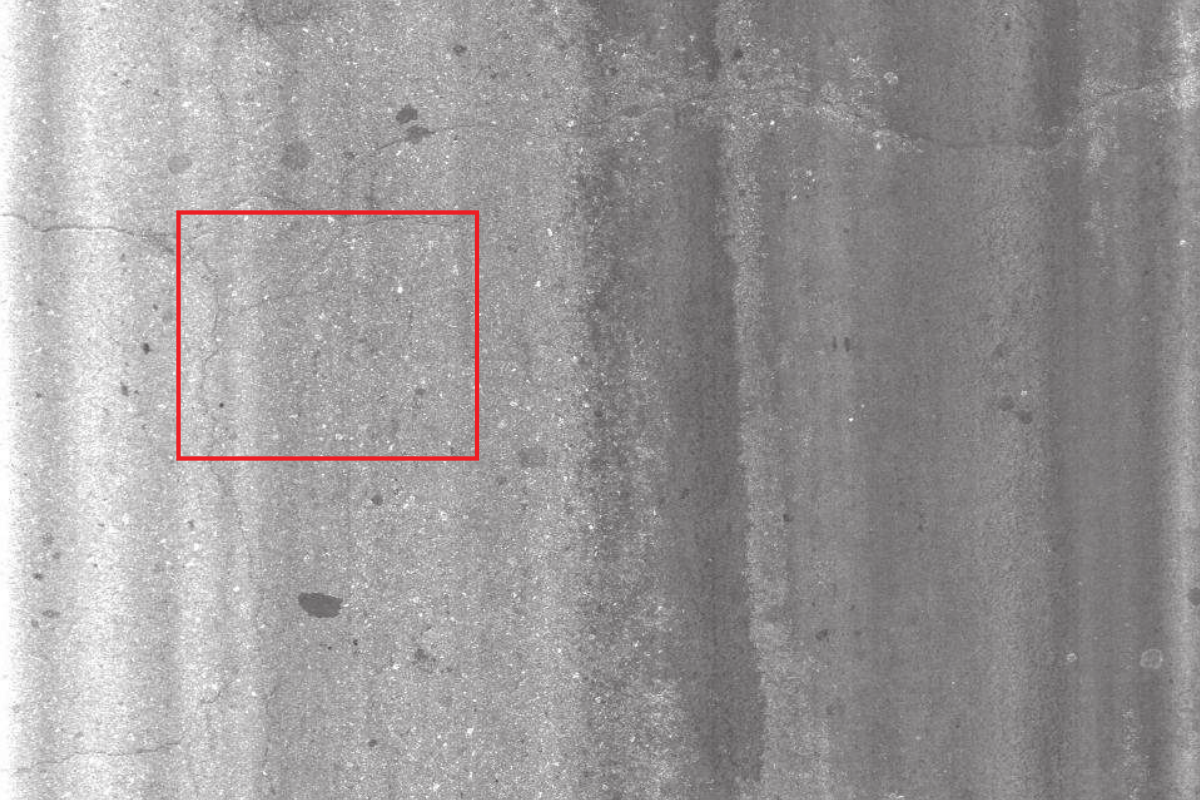}}
\subfloat[Result of our method]{\includegraphics[width = 0.333\textwidth]{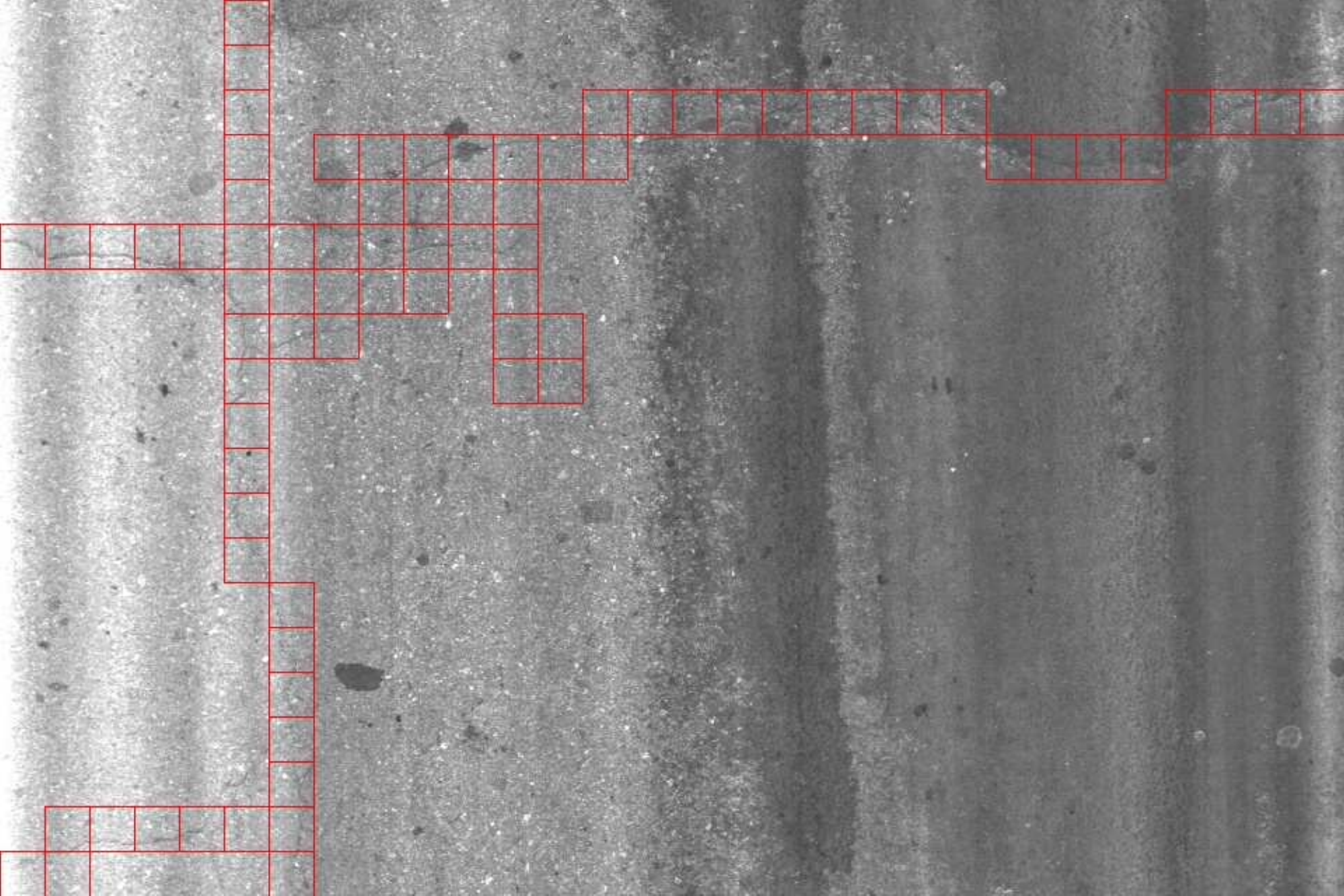}}
\subfloat[Binarized ground truth]{\includegraphics[width = 0.333\textwidth]{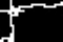}}
\caption{A result of MGCrackNet on BPC. Cracks in the red rectangle are difficult labeled in Fig.~\ref{img:image_data}(a), and please zoom out the image for the more detailed cracks.}
\label{img:image_data}
\end{center}
\end{figure*}

The core issue of crack detection is to achieve the precise crack localization and area. As shown in Fig.~\ref{img:image_data}, both the unique structure and the special appearance the cracks pose significant challenges. Specially, bitumen pavement as a mix of bitumen and gravels naturally forms a crack-like road surface.
Fig.~\ref{img:image_data} illustrates that the low contrast between cracks and background, weak continuity of cracks, and complex noise interference with cracks make a significant challenges to the practical crack detection tasks, especially for the bituminous pavement roads as follows:
\begin{enumerate}[]
 \item{\textbf{Low contrast}}: Small and unclear cracks have a low contrast with the background. Besides, the high similarity between the cracks and the background makes many cracks too difficult to be accurately labeled.
\item{\textbf{Weak continuity}}: The crack has a group of structures with a irregular and weak continuity distribution. It indicates that cracks would occur at any position on roads.
\item{\textbf{Noise interference}}: The background contains complex noises, such as the spots with different shapes and the scratches with varying lengths. These crack-like noises seriously affect the performance of crack detection methods.
\end{enumerate}

Both semantic segmentation~\cite{fang2022external}~\cite{ronneberger2015u} and object detection~\cite{xiang2022improved}~\cite{tsuchiya2019method} are the two mainstream methods for the crack detection task. However, due to the unique spatial structure of cracks, we find that both methods have their drawbacks. For the segmentation approach, pixel-level result guarantees the precise crack detection, but it requires the pixel-level annotations which cause a significant amount of time consumption. As shown in Fig.~\ref{img:image_data}~(a), cracks in the red rectangle are almost impossible to be accurately labeled pixel-level labeling. For the object detection method, it requires a bounding box to label the crack positions. However, the bounding box barely reflects the weak continuity of cracks, being unable to accurately label and measure the detailed cracks.

Compared to the semantic segmentation approach and the object detection one, as illustrated in Fig.~\ref{img:image_data}~(b), patch-level annotation is an approach to balance between efficiency and accuracy. That is, the patch-level annotation significantly reduces the difficulty of annotation. It seems that if an image patch is considered as a pixel, approaches from the semantic segmentation approach can readily be used for the crack detection. However, the patch-level annotation contains significant more noises than that of the pixel-level annotation, as illustrated in Fig.~\ref{img:image_data}~(b). These annotation makes the approaches for the semantic segmentation task inefficiently for crack detection. For instance, U-Net~\cite{ronneberger2015u} predicts multiple results by easily converting pixel-level annotation into the different scales. However, the patch-level annotation barely generates the accurate ground truths for the different scales for U-Net.

\begin{figure*}[t!]
\centerline{\includegraphics[width=6.5in, keepaspectratio]{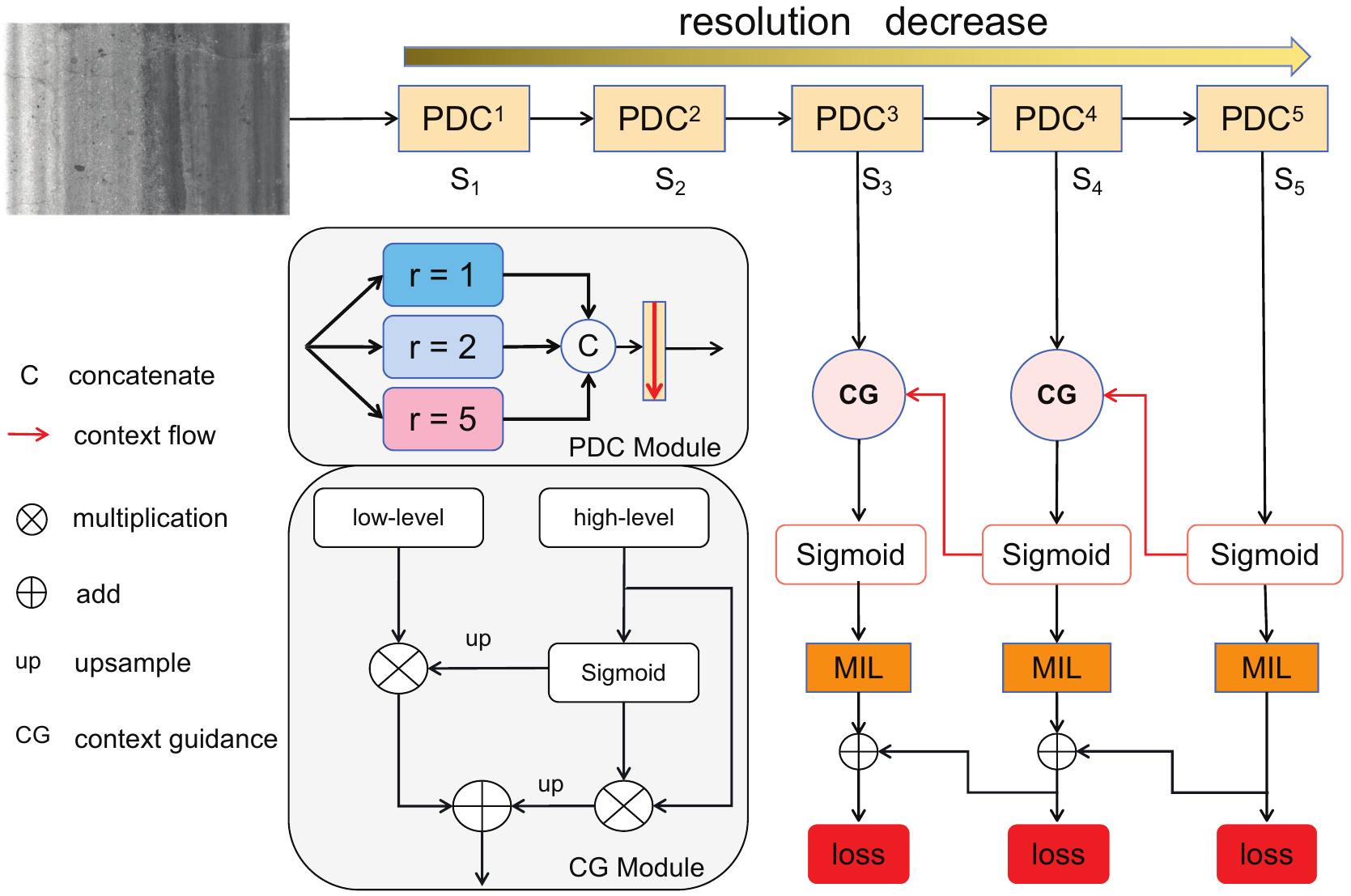}}
\caption{The network structure of MGCrackNet, which is composed of context feature extraction stages, context guidance module, MIL strategy and loss function.}
\label{img:model_structure}
\end{figure*}

In this paper, we propose to fuse \textbf{M}ulti \textbf{G}ranularity context information to model a \textbf{Crack} detection neural \textbf{Net}work (MGCrackNet), as illustrated in Fig.~\ref{img:model_structure}. To our best knowledge, MGCrackNet firstly leverages the dilated convolution to extract more local context information, where local features (\emph{e.g.} textural details of roads) and the spital information is captured to precisely locate cracks. Naturally, deep layers have more semantic information than that of the shallow ones~\cite{fan2020camouflaged}. Therefore, two granularity-level context are should be fused to model the context of cracks.

A natural idea is that the deep layers need the detailed spatial information to precisely detect cracks, while the shallow layers need to be guided by the semantic information of deep layers.
To achieve this bootstrap-like procedure, we firstly freeze shallow layers and only optimize features from the deep layers. The advantage of this step is that semantic information can be a guidance information for the shallow layers. Secondly, we unfroze and optimize the shallow layers, forcing the multi-granularity context information to gradually optimize the features from the deep layers to shallow ones. Besides, to align the context cues from the the deep layers to shallow ones, we introduce Multiple Instance Learning (MIL) strategy~\cite{8957372}\cite{wang2018revisiting}\cite{ilse2020deep} to further improve crack detection performance. Comparative experiments and ablation studies on three datasets demonstrate the proposed method achieves state-of-the-art performance, showing the superior ability of our network.

The contributions of this work are summarized in the two-fold:
\begin{itemize}
\item  To our best knowledge, to handle the patch-level annotation, we firstly propose multi-granularity context information flow to gradually guide context cues from deep layers to shallow ones. Besides, the bootstrap-like training procedure is proposed to efficiently fuse context cues and local ones.
\item To our best knowledge, we release the largest, the most complex and challenge bitumen pavement crack dataset, which adopts the patch-level annotation. It brings new and considerable challenges for the crack detection community.
\end{itemize}

\section{Related Work}\label{sec:relatedwork}

\textbf{Pixel-Level Crack Detection:} Pixel-level crack detection is a semantic segmentation based method that aims accurately to recognize the crack pixels. The key to crack pixels recognition lies in distinguishing between cracks and background. For example, the physical property is a straightforward way for crack detection including spatial continuity~\cite{liu2022asphalt}~\cite{han2021crackw}~\cite{choi2019sddnet}. Besides, fusing context information is an efficiency method~\cite{qu2021deeply}~\cite{xiang2022crack}~\cite{song2020automated}~\cite{liu2019deepcrack}. However, most of these cracks in the experiments are serious ones which are easily labeled cement pavement.

Due to the complex spatial structure of cracks, it is difficult to label cracks in a pixel wise approach. In practice, the industrial requirement pays attention to these mild yet unclear cracks, because the repair cost for these cracks is lower than that of these serious ones.

\textbf{Region-Level Crack Detection:} Region-level crack detection is an object detection based method that essentially judges whether there is a crack or not. For example, You Only Look Once v3 (YOLOv3) is combined with deformable convolution to locate cracks~\cite{tsuchiya2019method}\cite{liu2022novel}. The method based on YOLOv5 construct 12 different attention models for crack detection~\cite{yao2022detection}. In addition,~\cite{haciefendiouglu2022concrete} uses fast R-CNN to detect cracks in order to handling different weather conditions and lighting levels. ~\cite{tran2022two} first uses Mask R-CNN coarse locate cracks and second determines the severity of crack damage.

The performance of the region-level crack detection is continuously improved with the progress of the object detection community. However, the predicted results are too imprecise to quantify the area of cracks. In fact, the method~\cite{li2023automatic} just detects these serious cracks because they are easily be labeled.

\textbf{Patch-Level Crack Detection:} Patch level crack detection is a classification method that divides a crack image into multiple patches and further determines which patch is a crack. In~\cite{yusof2018crack}, a deep CNN is proposed to classify the image patches. In~\cite{feng2019multi}~\cite{Wang2022I2CNetAI}, context information from the different layers are used to improve the predicted results. In~\cite{duan2021pavement}, the optimization of crack characteristics at each stage improves the crack detection performance, which adopts the idea of modeling context information.

However, The above methods only rely on feature-level context and barely utilizes semantic contextual information. Besides, these methods ignore that the patch-level label is a weak annotation, which results in the inefficient feature extraction. Consequently, the context cues are inaccurate to efficiently improve detection performance. On the contrary, our Context Guidance (CG) module not only utilizes the context cues but also refines the feature extraction with the help of the MIL-based label alignment.

\section{Modeling Context Information}\label{section3}

\subsection{MGCrackNet}

In this paper, we adopt the parallel dilated convolution(PDC)~\cite{DBLP:journals/corr/WangCYLHHC17} module to extract local context of cracks. As shown in the PDC module in Fig.~\ref{img:model_structure}, the PDC module is composed of 3 dilated convolutional layers with the dilated rates of $r \in\{ 1, 2, 5\}$ as follows:
\begin{equation}\label{eqt:dilated}
O_{r} = C_{r}(F_i)
\end{equation}
where $F_i$ is the feature, $C_{r}(\cdot)$ is convolution operator with the dilated rate $r$. The dilated convolution~\eqref{eqt:dilated} inserts holes (zero) to expand the size of convolution kernel, extracting more context features than that of the standard convolution. However, these holes also result in the loss of information. Concatenate features $S_{o} = concat(O_{1}, O_{2}, O_{5})$ compensate for the loss of these holes in convolution.

\begin{figure}[t!]
\centering
    \centerline{\includegraphics[width=3.0in, keepaspectratio]{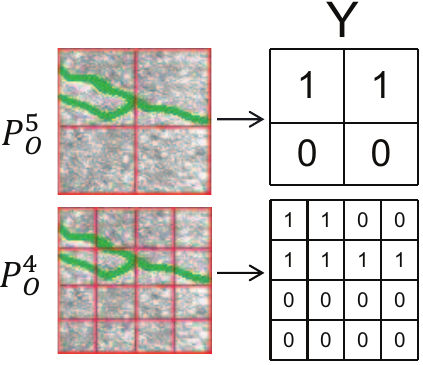}}
    \caption{A patch-level annotation across two consecutive stages.}
\label{fig:mil_patch}
\end{figure}

The PDC naturally obtains a larger context than that of the standard convolution. As shown in Fig.~\ref{fig:mil_patch}, in a patch with cracks, only a small pixels belong to cracks, while other regions is background or noises. Finally, we apply the max-pooling operation to sub-sampling the result of PDC~\eqref{eqt:dilated}. We stack PDC and max-pooling as the backbone of network, as shown in Fig.~\ref{img:model_structure}.

As shown in Fig.~\ref{img:model_structure}, MGCrackNet further sperate the backbone of network into different stages.
\begin{myDef}
	\label{def:stage}
    (Stage). The Stage is a repeatable feature extraction module in a neural network. Usually, a stage consists of the combination of the convolution-pooling operations.
\end{myDef}
In this paper, we split the network into 5 stages $ S_{i} $, $(i \in \{1,2,3,4,5\})$. Features extracted from $5$ stages are respectively donated as $ F_{S_{i}} \in \mathbb{R}^{{\frac{1}{2^{i}}H} \times \frac{1}{2^{i}}W \times C_{i} }$, $(C_{i} \in \{32,64,128,256,256\})$, $(i \in \{1,2,3,4,5\}) $, where $C_{i}$ is the number of channels for the $i$-th stage.

The previous approach~\cite{duan2021pavement} stacks several backbones of the network into a new network to boost the performance. However, stacking multiple backbones naturally lead to an over-load yet inefficient neural network during the training and the deployment stages. On the contrary, our stage-wise approach is more efficiently.

In this paper, we only use the feature maps of the last three stages, $ F_{S_{3}}\in \mathbb{R}^{{\frac{1}{8}H} \times \frac{1}{8}W \times C_{3} }$, $F_{S_{4}}\in \mathbb{R}^{{\frac{1}{16}H} \times \frac{1}{16}W \times C_{4} }$ and $F_{S_{5}}\in \mathbb{R}^{{\frac{1}{32}H} \times \frac{1}{32}W \times C_{5} }$. Because the last three stages contain less noise interference~\cite{fan2021concealed}. Concretely, $ F_{S_{i-1}}$ represents the local context of cracks, $ F_{S_{i}}$ contains the high-level semantic information, but also include more noise interference~\cite{mei2021camouflaged}. Therefore, a reasonable approach is to combine the advantages of both high-level features and the low-level ones.

Consequently, we propose to model context information flow between two stages, where the semantic context guides the extraction of the local context of cracks. During the guidance procedure, we use MIL to align the annotation between two consecutive stages. By this process, we gradually optimize the feature maps at each stage with the same loss functions.

\subsection{Context Guidance Module}
The CG module leverages the context cues to help the shallower stage extract informative features and local context, as illustrated in Fig.~\ref{img:model_structure}. Concretely, the deeper stage generates a heatmap of cracks, where each value in the map means the probability of whether a patch contains a crack or not. Denote the features extracted by the stage $S_{i}$ as $F_{S_{i}}$, the corresponding feature map $M_{S_{i}}$ is computed as follows:
\begin{equation}\label{eqt:attention-mask}
M_{S_{i}} = \sigma (F_{S_{i}})
\end{equation}
where $\sigma (\cdot)$ is the sigmoid function, index $ i \in \{3,4,5\} $ and $m_{xy} \in M_{S_{i}}$ is the probability weights, $ M_{S_{i}}\in \mathbb{R}^{{H_{S_{i}}} \times W_{S_{i}} \times C_{i}}$, $(H_{S_{i}} = {\frac{1}{2^{i}}H}, W_{S_{i}} = {\frac{1}{2^{i}}W})$, $m_{xy} \in [0,1]$, $x=1,\ldots,H_{S_{i}}, y=1,\ldots,W_{S_{i}}$, where $H_{S_{i}}$ and $W_{S_{i}}$ are the number of the patches in a image.

The heatmap $M_{S_{i}}$ acts as an attention mechanise to guide the feature extraction of the shallow layer, assigning a higher value to the crack position. Besides, the features $F_{S_{i}}$ contain the context cues of the consecutive stage, because the receptive field of the stage $S_{i}$ is larger than the stage $S_{i-1}$. Therefore, we further enhance the discriminative ability of feature $F_{S_{i}}$ as follows:
\begin{equation}
\hat{F}_{S_{i}} = M_{S_{i}} \otimes F_{S_{i}}
\end{equation}
where $\otimes$ is the element-wise multiplication. Thus, the weighted feature $\hat{F}_{S_{i}}$ is a self-refined feature.

To align the feature size between two consecutive stages, the heatmap $M_{S_{i}}$ is upsampled through bilinear interpolation operation. The low-level feature $F_{S_{i-1}}$ is weighted by the updampled heatmap $up(M_{S_{i}})$ as follows:
\begin{equation}\label{eqt:contextguided-feature}
\hat{F}_{S_{i-1}} = up(M_{S_{i}}) \otimes F_{S_{i-1}}
\end{equation}
where $up(\cdot)$ is the upsampling function.

If we consider $M_{S_{i}}$ as the context of the stage $S_{i-1}$, $\hat{F}_{S_{i-1}}$ is the context guided features. Concretely, in~\eqref{eqt:contextguided-feature}, the element-wise multiplication guides feature $F_{S_{i-1}}$ to pay attention to the important positions in the receptive field of the shallow stage. To verify this assumption, the gradient of feature$\hat{F}_{S_{i-1}}$ with respect to parameters $\theta$ as follows:
\begin{equation}\label{eqt:grad-CG-feature}
\frac{\partial \hat{F}_{S_{i-1}}}{\partial \theta} = \frac{\partial up(M_{S_{i}})}{\partial\theta} \otimes F_{S_{i-1}} + up(M_{S_{i}}) \otimes \frac{\partial  F_{S_{i-1}}}{\partial \theta}
\end{equation}
where $\frac{\partial \cdot }{\partial \theta}$ is the gradient of the different features with respect to parameters $\theta$.

Obviously, in~\eqref{eqt:grad-CG-feature}, the gradient $\frac{\partial \hat{F}_{S_{i-1}}}{\partial \theta}$ is controlled by four terms, \emph{e.g.} $M_{S_{i}}$, $\frac{\partial up(M_{S_{i}})}{\partial\theta}$, $F_{S_{i-1}}$, and $\frac{\partial  F_{S_{i-1}}}{\partial \theta}$, in which the heatmap $M_{S_{i}}$ is generated by the feature $F_{S_{i-1}}$ from the previous stage $S_{i-1}$. If we simultaneously optimize $F_{S_{i-1}}$ and $M_{S_{i}}$, a possible gradient contradicts would happen between $F_{S_{i-1}}$ and $M_{S_{i}}$. Consequently, the inaccurate heatmap $M_{S_{i}}$ tends to fail guide the features for the shallow stages. Therefore, a natural yet effective strategy is firstly to froze $ F_{S_{i-1}}$ and optimize $M_{S_{i}}$ as follows:
\begin{equation}\label{eqt:frozen-grad-CG-feature}
\frac{\partial \hat{F}_{S_{i-1}}}{\partial \theta} = \frac{\partial up(M_{S_{i}})}{\partial\theta} \otimes F_{S_{i-1}}
\end{equation}

To retain the sufficient priors from the deeper stage and leverage the features from the shallow one, the upsampled feature $up(\hat{F}_{S_{i}})$ is element-wisely added to the features $\hat{F}_{S_{i-1}}$ from the shallow stage as follows:
\begin{equation}
F_{S_{i}}^{o} = add(up(\hat{F}_{S_{i}}), \hat{F}_{S_{i-1}})
\end{equation}
where $F_{S_{i}}^{o}$ is the weighted low-level feature at the stage $S_{i}$, and $add(\cdot,\cdot)$ is the element-wise addition operation.

\begin{figure*}[!t]\scriptsize
\begin{center}

\subfloat{\includegraphics[width=0.28\textwidth]{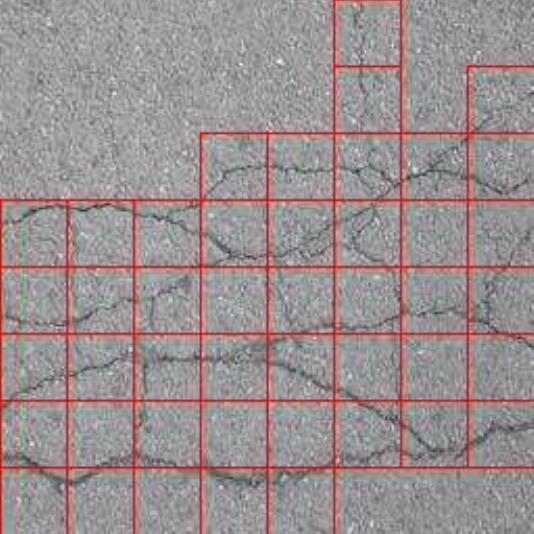}}\hspace{1mm}
\subfloat{\includegraphics[width=0.281\textwidth]{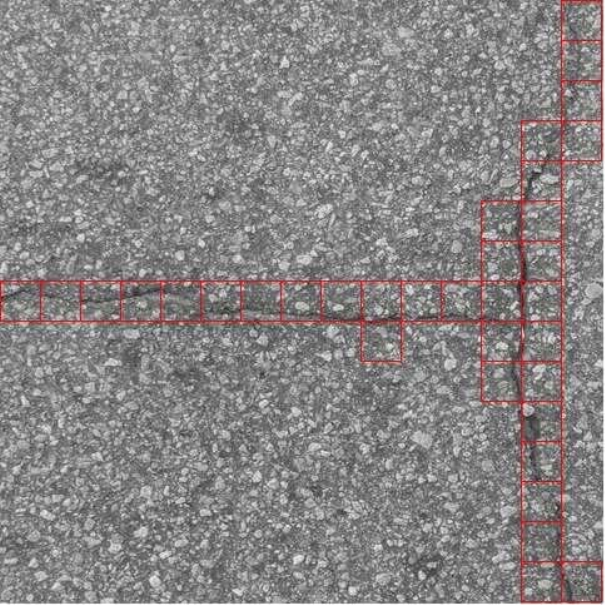}}\hspace{1mm}
\subfloat{\includegraphics[width=0.28\textwidth]{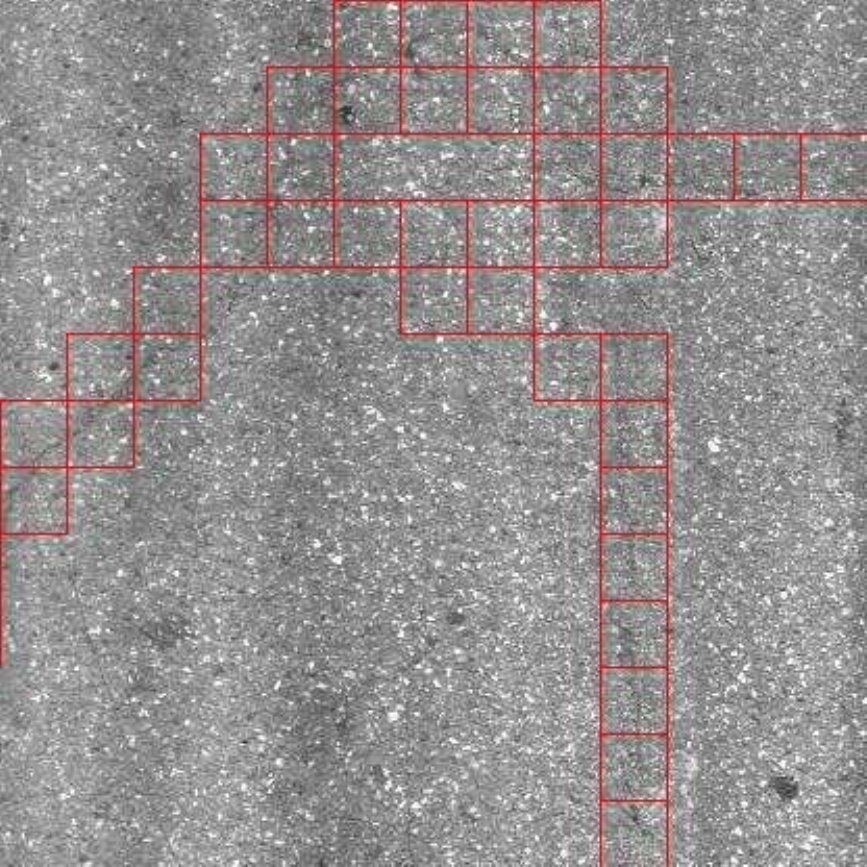}}

\subfloat{\includegraphics[width=0.28\textwidth]{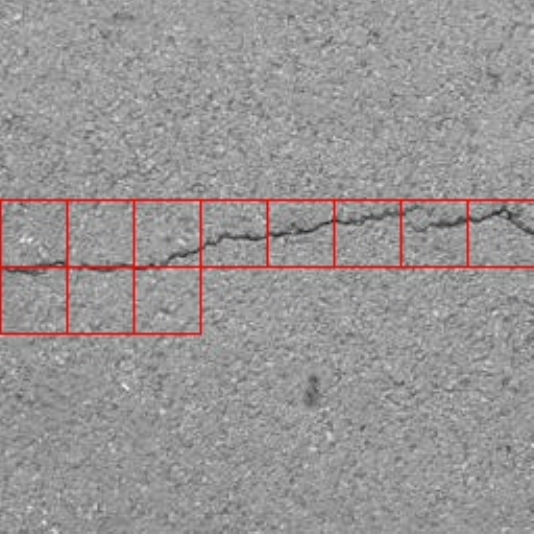}}\hspace{1mm}
\subfloat{\includegraphics[width=0.281\textwidth]{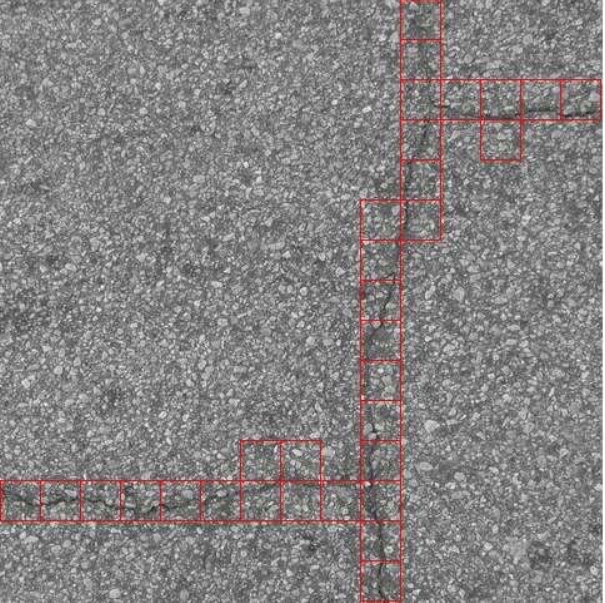}}\hspace{1mm}
\subfloat{\includegraphics[width=0.28\textwidth]{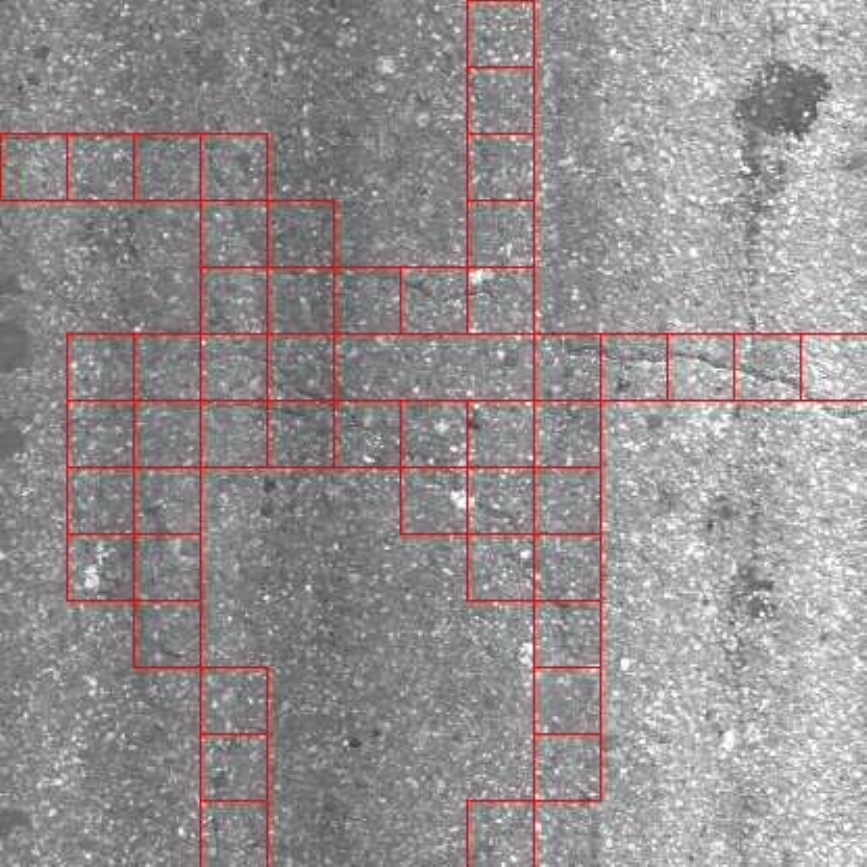}}

\setcounter {subfigure} {0}
\subfloat[CFD]{\includegraphics[width=0.28\textwidth]{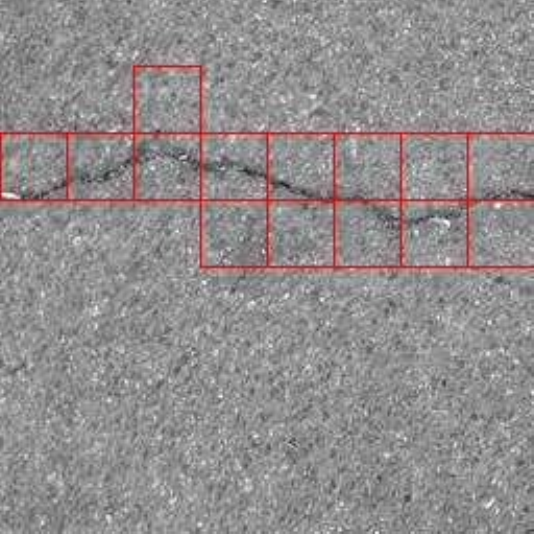}}\hspace{1mm}
\subfloat[Cracktree]{\includegraphics[width=0.281\textwidth]{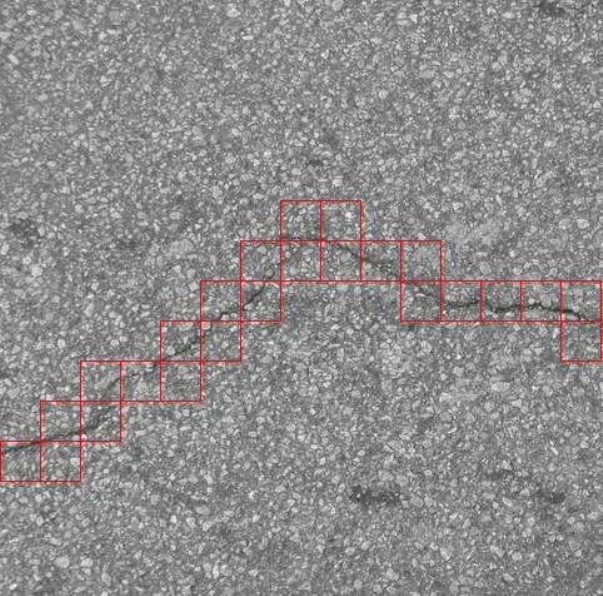}}\hspace{1mm}
\subfloat[BPC]{\includegraphics[width=0.28\textwidth]{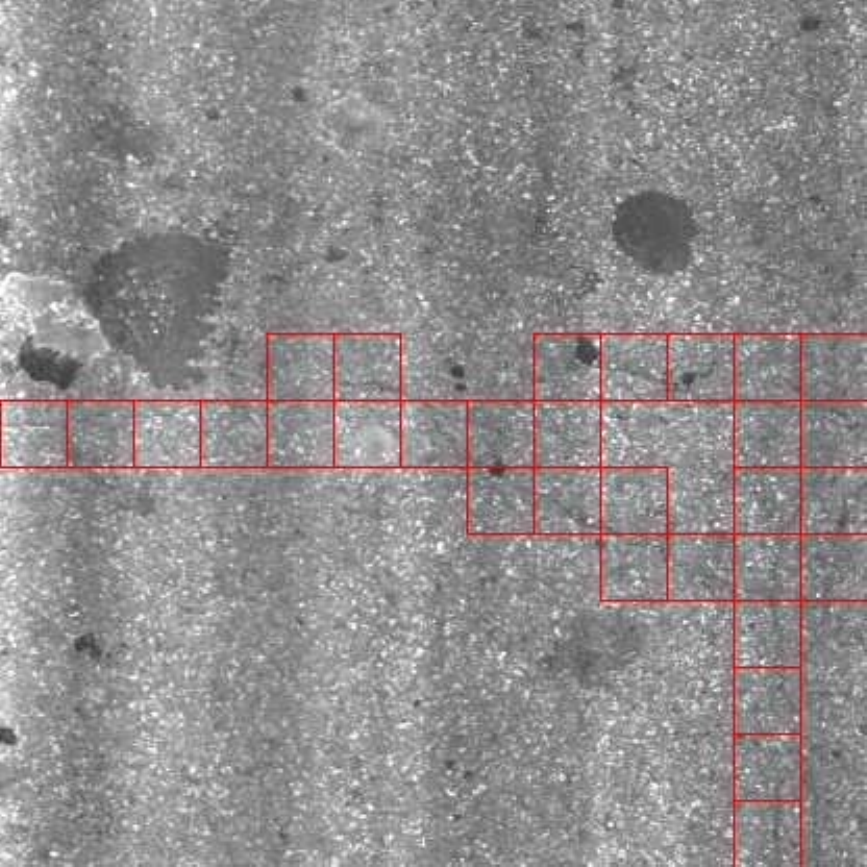}}

\caption{Some samples from three datasets. The columns from the left to the right represent the ground truth of CFD, Cracktree and BPC dataset, respectively.}
\label{fig:dataset_samples}
\end{center}
\end{figure*}

\subsection{Label Alignment for Stage-Wise Loss} An naive idea is that if we supervise network at the different stages, we will extract more discriminative features. However, the label alignment problem naturally happens, once we supervise the MGCrackNet at the different stages. Because the ground truth is annotated at the patch-wise granularity rather than the pixel-wise one. Consequently, we can not directly compute a loss at different stages. For example, as shown in Fig.~\ref{fig:mil_patch}, an $128\times 128$ image is divided into $4\times 4$ patches, where each $32\times 32$ patch is labeled whether it contains a crack or not for the stage $S_5$. Unfortunately, the annotation of the $16\times 16$ sub-patches cannot generated from the $4\times 4$ patch-wise annotation. we call this problem as the label alignment problem.

We apply the MIL strategy to solve the label alignment problem. Concretely, a ``bag'' corresponds to a patch, while an ``instance'' corresponds to a sub-patch in a patch. A classifier is firstly built by a $3 \times 3$ standard convolution and two $1 \times1 $ standard convolutions for each sub-regions as follows:
\begin{equation}
P_{o}^{i} = \sigma (F^{o}_{S_{i}})
\end{equation}
where $\sigma(\cdot)$ is the sigmoid function and $P_{o}^{i}$ is the classification probability, $ P_{o}^{i} \in \mathbb{R}^{{\frac{1}{2^{i}} H} \times \frac{1}{2^{i}} W \times C_{i}}$ for the stage $S_i$.

Secondly, MIL strategy is used to convert the probability $P_{o}^{i}$ of the sub-regions to the patch-wise annotation as follows:
\begin{equation}
Y_{o}^{i} = pool(P_{o}^{i})
\end{equation}
where $pool(\cdot)$ is the pooling function, $Y_{o}^{i}$ is the classification probability at the different stages, $Y_{o}^{i} \in \mathbb{R}^{{\frac{1}{32} H} \times \frac{1}{32} W \times C }$. In MGCrackNet, the function $pool(\cdot)$ uses stride \{4, 2, 1\} for the $S_{i}$, respectively. During inference, if the probability $Y_{o}^{i}$ is greater than $0.5$, it indicates that a patch contains cracks.

\subsection{Loss function} We employ a multi-stage supervised approach to optimize the predicted results, as shown in Fig.~\ref{img:model_structure}. Concretely, given a labeled sample $\{X,Y\}$, where $X \in \mathbb{R}^{{\frac{1}{2^i} H} \times \frac{1}{2^i} W \times C }$ is an image, $Y\in \mathbb{R}^{{\frac{1}{32} H} \times \frac{1}{32} W \times C }$ is the corresponding multi-stage label, and $i \in \{3,4,5\}$ are index of the different stages, Cross Entropy (CE) loss of a sample $\{X,Y\}$ is used at different stages as follows:
\begin{equation}
L = \sum_{i = 3}^{5}  CE(Y_{o}^{i},Y)
\end{equation}
where $CE(\cdot ,\cdot )$ is the cross entropy loss.

\section{Experiments}\label{sec:section4}

\subsection{Dataset Comparisons}

1) \textbf{CFD:} The CFD~\cite{shi2016automatic} dataset is for the cement road crack. CFD contains 118 crack images with a size of $480\times320$. We use the $256\times256$ sliding window to crop the images, generating $2,832$ samples, where $2,256$ samples are used for the training and $576$ ones are for the testing. Each sample is divided into $32\times32$ patches. Some samples from CFD are shown in Fig.~\ref{fig:dataset_samples}~(a).

2) \textbf{Cracktree:} The Cracktree dataset is also for the cement road crack~\cite{zou2012cracktree}. This dataset consists of $206$ crack images with a size of $800\times600$. We use the $512\times512$ sliding window to generate the $1,300$ training samples and the $324$ testing ones. Each samples is further divided into $32\times32$ patches. Some samples from Cracktree are shown in Fig.~\ref{fig:dataset_samples}~(b).

3) \textbf{BPC:} We release the challengeable Bitumen Pavement Crack (BPC) dataset, which comprises $3,522$ images with a size of $640\times 960$. Each image is divided into $16\times 16$ patches with a size of $32\times 32$. We use the $416\times 416$  sliding window to generate $7,044$ samples, where the $5,764$ training samples and the $1,280$ testing ones. Some samples from BPC are shown in Fig.~\ref{fig:dataset_samples}~(c). Rather than Cracktree and CFD, BPC contains about $3,080$ noisy patches which are difficult to be correctly labeled by annotators. Because this problem is caused by the materials (\emph{e.g.} the mix of bitumen and gravel) of the bitumen pavement itself.

Due to the complexity of bitumen pavement cracks, the cracks and background are highly similar, some of these patches barely be subjectively judged by humans (see Fig.~\ref{results_visualization}). During the labeling process, we have investigated the experts to label these patches. These cracks make BPC challenge to the SOTA methods.

\begin{figure}[!t]\scriptsize
\begin{center}
\subfloat[pixel-level]{\includegraphics[width=0.23\textwidth]{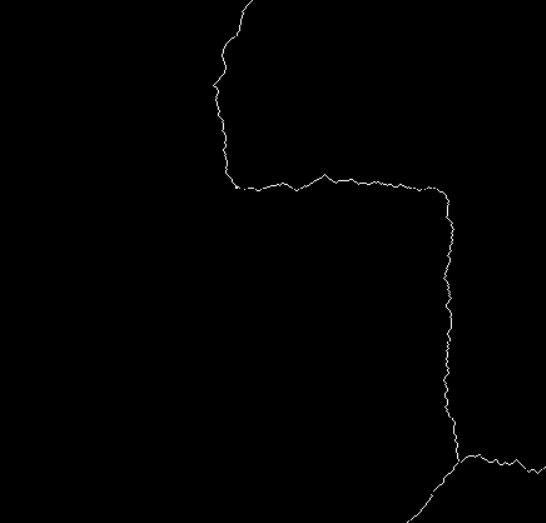}}\hspace{1mm}
\subfloat[patch-level]{\includegraphics[width=0.22\textwidth]{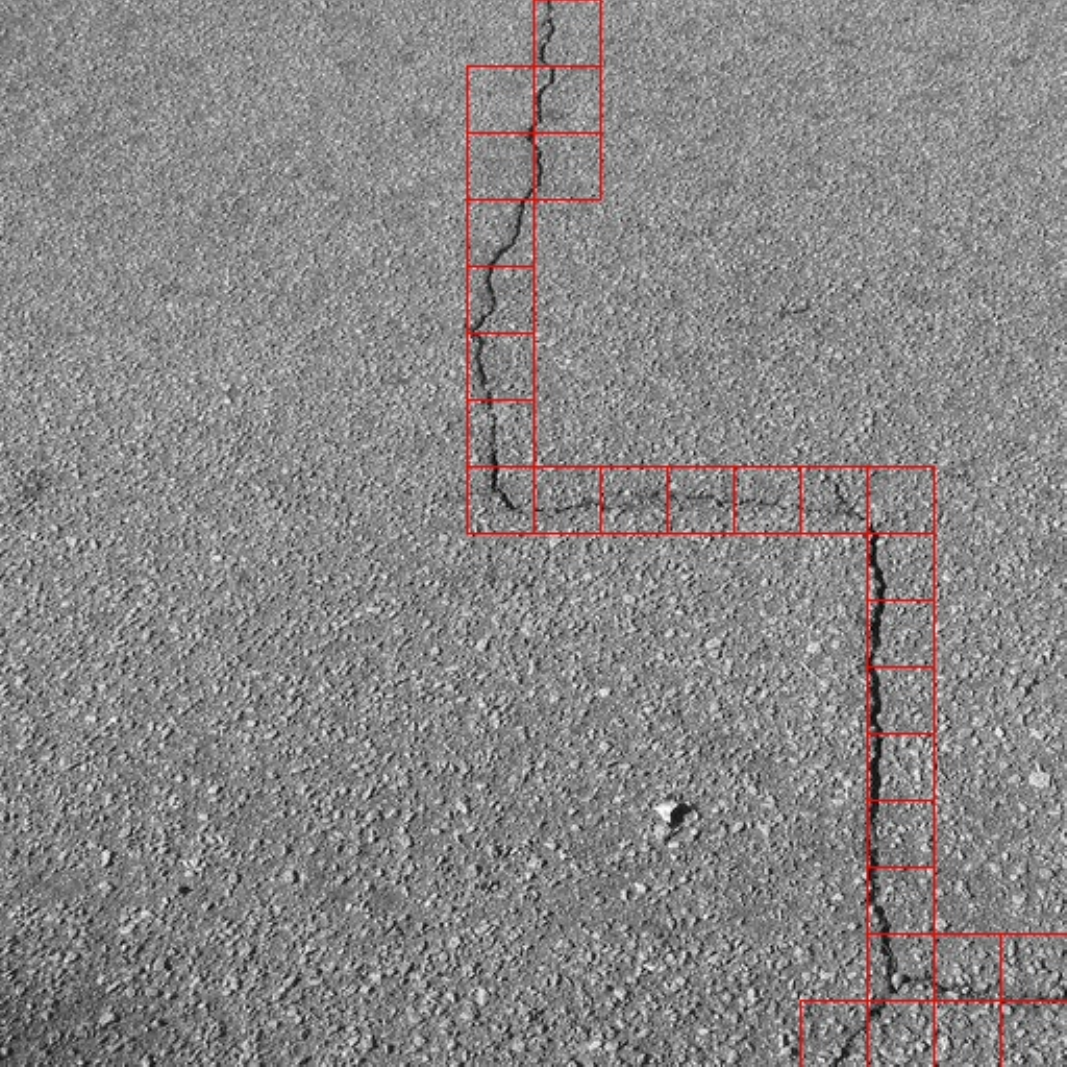}}
\caption{The pixel-level annotation is converted into the the patch-level one.}
\label{img:pixeltopatch}
\end{center}
\end{figure}

It should note that both CFD and Cracktree are the pixel-level annotation, which does not meet our setting. During experiments, we regenerate patch-level labels from the pixel-level annotation as shown in Fig.~\ref{img:pixeltopatch}.

\begin{table}[t!]
\centering
\renewcommand\arraystretch{1.5}
\tabcolsep=0.45cm
\caption{Comparison Between Three Datasets}
\begin{tabular}{|c|c|c|c|}
\hline
Attributes                              & BPC             & CFD         & Cracktree  \\ \hline
Road Types                              & Bitumen         & Cement      & Cement       \\
Image Size                              & $640\times 960$?         & 480x320     & 800x600      \\
Patch Size                              & $32x32$           & 32x32       & 32x32        \\
Sliding Window                          & 416x416         & 256x256     & 512x512      \\
\hline
\# Image                                & 5764            & 2832        & 1624         \\
\# Noise Patches                        & 3080            & 0           & 0          \\
$\frac{\# Bg Patch}{\# Crack Patch}$    & 16 :1             & 3 :1          & 6 :1         \\
\hline
\end{tabular}
\label{dataset_information}
\end{table}

As compared with both public datasets, the purpose that we release BPC for the research community are two folds:
\begin{itemize}
\item The cracks of bitumen pavement are more changeable than that of the cement road. Our experimental results also confirm this observation.
\item The patch-level crack detection, a weakly supervised approach, is a challenge yet practical problem. Reducing the annotation difficulty yet obtaining a satisfied accurate performance is a practical industrial requirement.
\end{itemize}

\subsection{Evaluation Metrics}

In this paper, $Precision(P)$ and $Recall(R)$ are adopted to evaluate the performances among different methods. The $F1$ score measures the performance from $Precision$ and $Recall$ as follows:
\begin{equation}
F1= 2\tfrac{P\cdot R}{P+R}
\end{equation}
where $P$ is $Precision$, $R$ is $Recall$.

To further evaluate the performance of crack detection, $Average Precision (AP)$ of the cracks is computed as a comprehensive metric to reflect the overall stability of the model as follows:
\begin{equation}
AP = \int_{0}^{1} P(t)dt
\end{equation}
where value $P(t)$ is the precision $P$ at the threshold $t$, $t\in [0,1]$. Therefore, $AP$ is equivalent to the area under the precision-recall curve, measures the performance of cracks rather than the background patches.

\begin{algorithm}[H]
\caption{Freeze Training Strategy.}\label{alg:alg1}
\begin{algorithmic}
\STATE
\STATE  \textbf{INPUT:} multi-stage features $\left \{{F_{S_{i}}, i \in {3,4,5}}\right \} $
\STATE  \hspace{1.2cm} iteration number $t=1,\ldots, N$
\STATE  \textbf{OUTPUT:} feature map $\left \{{Y_{o}^{i}, i\in {3,4,5}}\right \}$
\STATE  ${Y_{o}^{i}|i = 3,4,5} \longleftarrow {F_{S_{i}}, i\in {3,4,5}}$;
\STATE  \textbf{for} $t = 1; t < N; t \longleftarrow t+1$ do
\STATE  \hspace{0.3cm}if $ 1 <= t <= 20 $ do
\STATE  \hspace{0.7cm}   $Y_{o}^{3}, Y_{o}^{4}, Y_{o}^{5} \longleftarrow freeze(F_{S_{3}}), freeze(F_{S_{4}}), CG(F_{S_{5}})$;
\STATE  \hspace{0.3cm}if $ 20 <t <= 40 $ do
\STATE  \hspace{0.7cm}   $Y_{o}^{3}, Y_{o}^{4}, Y_{o}^{5} \longleftarrow freeze(F_{S_{3}}), CG(F_{S_{4}}), CG(F_{S_{5}})$;
\STATE  \hspace{0.3cm}if $40 < t <= N $ do
\STATE  \hspace{0.7cm}   $Y_{o}^{3}, Y_{o}^{4}, Y_{o}^{5} \longleftarrow CG(F_{S_{3}}), CG(F_{S_{4}}), CG(F_{S_{5}})$;
\STATE  \textbf{end}
\STATE  \textbf{return}  $\left \{{Y_{o}^{i}}\right \}$;
\end{algorithmic}
\label{alg1}
\end{algorithm}

\begin{table*}[]
\centering
\caption{Contrasting Experimental Results on Three Datasets}
\begin{tabular}{c|cccc|cccc|cccc}
\hline
\multicolumn{1}{l||}{\multirow{2}{*}{Methods}} & \multicolumn{4}{c||}{BPC} & \multicolumn{4}{c||}{CFD}&\multicolumn{4}{c}{Cracktree} \\ \cline{2-13}
\multicolumn{1}{l||}{\multirow{2}{*}{}}
& \multicolumn{1}{c|}{$P$($\%$)}  & \multicolumn{1}{c|}{$R$($\%$)} & \multicolumn{1}{c|}{$F1$($\%$)}  & \multicolumn{1}{c||}{$AP$($\%$)}
& \multicolumn{1}{c|}{$P$($\%$)}  & \multicolumn{1}{c|}{$R$($\%$)} & \multicolumn{1}{c|}{$F1$($\%$)}  & \multicolumn{1}{c||}{$AP$($\%$)}
& \multicolumn{1}{c|}{$P$($\%$)}  & \multicolumn{1}{c|}{$R$($\%$)} & \multicolumn{1}{c|}{$F1$($\%$)}  & $AP$($\%$)              \\ \hline
\multicolumn{1}{l||}{ST \cite{liu2021swin}}
& \multicolumn{1}{c|}{0.00}  & \multicolumn{1}{c|}{0.00} & \multicolumn{1}{c|}{0.00}    & \multicolumn{1}{c||}{9.16}
& \multicolumn{1}{c|}{86.90} & \multicolumn{1}{c|}{72.56} & \multicolumn{1}{c|}{79.08}  & \multicolumn{1}{c||}{86.98}
& \multicolumn{1}{c|}{85.66} & \multicolumn{1}{c|}{85.58}  & \multicolumn{1}{c|}{85.61}  & 90.09           \\
\multicolumn{1}{l||}{FT \cite{yang2021focal}}
& \multicolumn{1}{c|}{0.00} & \multicolumn{1}{c|}{0.00} & \multicolumn{1}{c|}{0.00}            & \multicolumn{1}{c||}{9.25}
& \multicolumn{1}{c|}{80.30} & \multicolumn{1}{c|}{67.01} & \multicolumn{1}{c|}{73.05}           & \multicolumn{1}{c||}{81.92}
& \multicolumn{1}{c|}{78.03} & \multicolumn{1}{c|}{75.42} & \multicolumn{1}{c|}{76.70}           & 84.23          \\ \hline\hline
\multicolumn{1}{l||}{DenseNet121 \cite{huang2017densely}}
& \multicolumn{1}{c|}{53.86} & \multicolumn{1}{c|}{64.64} & \multicolumn{1}{c|}{58.75}          & \multicolumn{1}{c||}{61.46}
& \multicolumn{1}{c|}{91.20} & \multicolumn{1}{c|}{79.60} & \multicolumn{1}{c|}{85.00}          & \multicolumn{1}{c||}{90.05}
& \multicolumn{1}{c|}{89.26} & \multicolumn{1}{c|}{87.22} & \multicolumn{1}{c|}{88.22}          & 93.67          \\
\multicolumn{1}{l||}{ResNet50 \cite{he2016deep}}
& \multicolumn{1}{c|}{76.70} & \multicolumn{1}{c|}{78.69} & \multicolumn{1}{c|}{77.68}          & \multicolumn{1}{c||}{79.41}
& \multicolumn{1}{c|}{94.68}  & \multicolumn{1}{c|}{84.53} & \multicolumn{1}{c|}{89.31}          & \multicolumn{1}{c||}{93.64}
& \multicolumn{1}{c|}{93.34}    & \multicolumn{1}{c|}{83.97} & \multicolumn{1}{c|}{89.55}          & 93.23          \\
\multicolumn{1}{l||}{VGG16 \cite{simonyan2014very}}
& \multicolumn{1}{c|}{77.95}  & \multicolumn{1}{c|}{81.40}  & \multicolumn{1}{c|}{\textbf{79.63}} & \multicolumn{1}{c||}{82.51}
& \multicolumn{1}{c|}{95.57}   & \multicolumn{1}{c|}{83.77}  & \multicolumn{1}{c|}{{89.28}} & \multicolumn{1}{c||}{94.76}
& \multicolumn{1}{c|}{\textbf{94.91}}   & \multicolumn{1}{c|}{84.77}  & \multicolumn{1}{c|}{{89.55}} & 93.23          \\ \hline\hline
\multicolumn{1}{l||}{DCNN \cite{yusof2018crack}}
& \multicolumn{1}{c|}{74.46} & \multicolumn{1}{c|}{68.44} & \multicolumn{1}{c|}{71.32}          & \multicolumn{1}{c||}{75.84}
& \multicolumn{1}{c|}{94.32}  & \multicolumn{1}{c|}{85.13}  & \multicolumn{1}{c|}{89.48}          & \multicolumn{1}{c||}{94.83}
& \multicolumn{1}{c|}{91.51}  & \multicolumn{1}{c|}{79.01}  & \multicolumn{1}{c|}{84.80}          & 91.94          \\
\multicolumn{1}{l||}{MSCNN \cite{feng2019multi}}
& \multicolumn{1}{c|}{77.58} & \multicolumn{1}{c|}{72.85} & \multicolumn{1}{c|}{75.14}          & \multicolumn{1}{c||}{80.20}
& \multicolumn{1}{c|}{96.93}  & \multicolumn{1}{c|}{85.52} & \multicolumn{1}{c|}{90.91}          & \multicolumn{1}{c||}{96.57}
& \multicolumn{1}{c|}{92.93}  & \multicolumn{1}{c|}{83.97}  & \multicolumn{1}{c|}{88.22}          & 93.85          \\
\multicolumn{1}{l||}{UFE \cite{pang2022you}}
& \multicolumn{1}{c|}{76.23}  & \multicolumn{1}{c|}{\textbf{80.17}}  & \multicolumn{1}{c|}{78.15}          & \multicolumn{1}{c||}{82.37}
& \multicolumn{1}{c|}{96.96}  & \multicolumn{1}{c|}{90.09}  & \multicolumn{1}{c|}{93.39}          & \multicolumn{1}{c||}{95.76}
& \multicolumn{1}{c|}{92.44}  & \multicolumn{1}{c|}{82.76}  & \multicolumn{1}{c|}{87.33}          & 92.63            \\
\multicolumn{1}{l||}{SFE \cite{duan2021pavement}}
& \multicolumn{1}{c|}{84.71} & \multicolumn{1}{c|}{73.35}  & \multicolumn{1}{c|}{78.62}   & \multicolumn{1}{c||}{85.10}
& \multicolumn{1}{c|}{{96.17}} & \multicolumn{1}{c|}{85.82}  & \multicolumn{1}{c|}{90.70}           & \multicolumn{1}{c||}{95.62}
& \multicolumn{1}{c|}{{92.55}} & \multicolumn{1}{c|}{89.93}  & \multicolumn{1}{c|}{\textbf{91.22}}    & 92.99          \\ \hline
\multicolumn{1}{l||}{MGCrackNet}
& \multicolumn{1}{c|}{\textbf{86.90}}  & \multicolumn{1}{c|}{72.45} & \multicolumn{1}{c|}{79.02} & \multicolumn{1}{c||}{\textbf{88.32}}
& \multicolumn{1}{c|}{\textbf{97.21}}  & \multicolumn{1}{c|}{\textbf{90.35}} & \multicolumn{1}{c|}{\textbf{93.65}} & \multicolumn{1}{c||}{\textbf{97.29}}
& \multicolumn{1}{c|}{90.21}  & \multicolumn{1}{c|}{\textbf{90.17}} & \multicolumn{1}{c|}{90.19} & \textbf{95.24} \\ \hline
\end{tabular}
\label{experiment-result}
\end{table*}

\subsection{Experiment Settings}

In our experiments, we use the open-source deep learning framework PyTorch and conduct the tests on NVIDIA 3080 GPUs. During training, we do not use any pre-trained models instead train the models from the scratch. We set the batch size as $16$ and use Stochastic Gradient Descent (SGD). Weight decay and momentum are set to $0.0005$ and 0.9, respectively. We set the initial learning rate to $1e-3$. After 40 epoches, it will be divided by 10 for every 20 epoches. we use data augmentation techniques such as random rotation, random cropping, and symmetric transformations to improve the model robustness.

In addition, we employe the freezing training strategy in Alg.~\ref{alg1}. Concretely, we freeze low-level features in the first 20 epoches and only optimize high-level features, and then open lower-level features in turn for training every 20 epoches.

\subsection{Baselines and The State-of-Art Approaches}

To demonstrate the superiority of the proposed method, we compare MGCrackNet with different methods on these public datasets from three aspects as follows:
\begin{itemize}
\item \textbf{Global Semantic Context}: Two general vision transformer networks, Swin Transformer~\cite{liu2021swin} (ST) and Focal Transformer~\cite{yang2021focal} (FT), are used as baseline to extract the global context features. Based on the Transformer~\cite{dosovitskiy2021image} architecture, ST and FT extract global context features from samples and achieve the best performance in visual classification tasks. Therefore, we use both methods as baselines to extract global semantic context features for crack detection.
\item\textbf{Local Context}: Three efficient architectures, VGG16~\cite{simonyan2014very}, ResNet50~\cite{he2016deep} and DenseNet121~\cite{huang2017densely}, are used as baseline to extract local context features. These algorithms are also successful in visual classification tasks.
\item \textbf{The State of Art Crack Detection Methods}: The State-Of-The-Art (SOTA) methods includes Deep Convolution Neural Network (DCNN)~\cite{yusof2018crack}, Multi-Scale classification network (MSCNN)~\cite{feng2019multi}, Structural Feature Extraction model(SFE)~\cite{duan2021pavement}, and Unconventional Feature Extraction model (UFE)~\cite{pang2022you}. These methods belong to patch-level crack detection algorithms. The reasons that we list them as SOTA methods as follows:
    \begin{itemize}
    \item \textbf{DCNN:} This paper detected cracks by a deep convolution network. However, this method ignores the importance of context information.
    \item \textbf{UFE and SFE:} Both UFE and SFE extract local context information by feature fusing strategy. However, these methods did not handle label alignment problem.
    \item \textbf{MSCNN:} This paper uses multiple max-pooling operations at different layers of the network to sub-sample the local feature. However, these max-pooling operations cause serious information loss.
    \end{itemize}
    For a fair comparison, the learning rate and the other hyper-parameters of these comparison methods are tuned to obtain the best performances for these STOA methods.
\end{itemize}

\begin{figure*}[t]
\begin{center}

\subfloat{
\includegraphics[width = 0.16\textwidth]{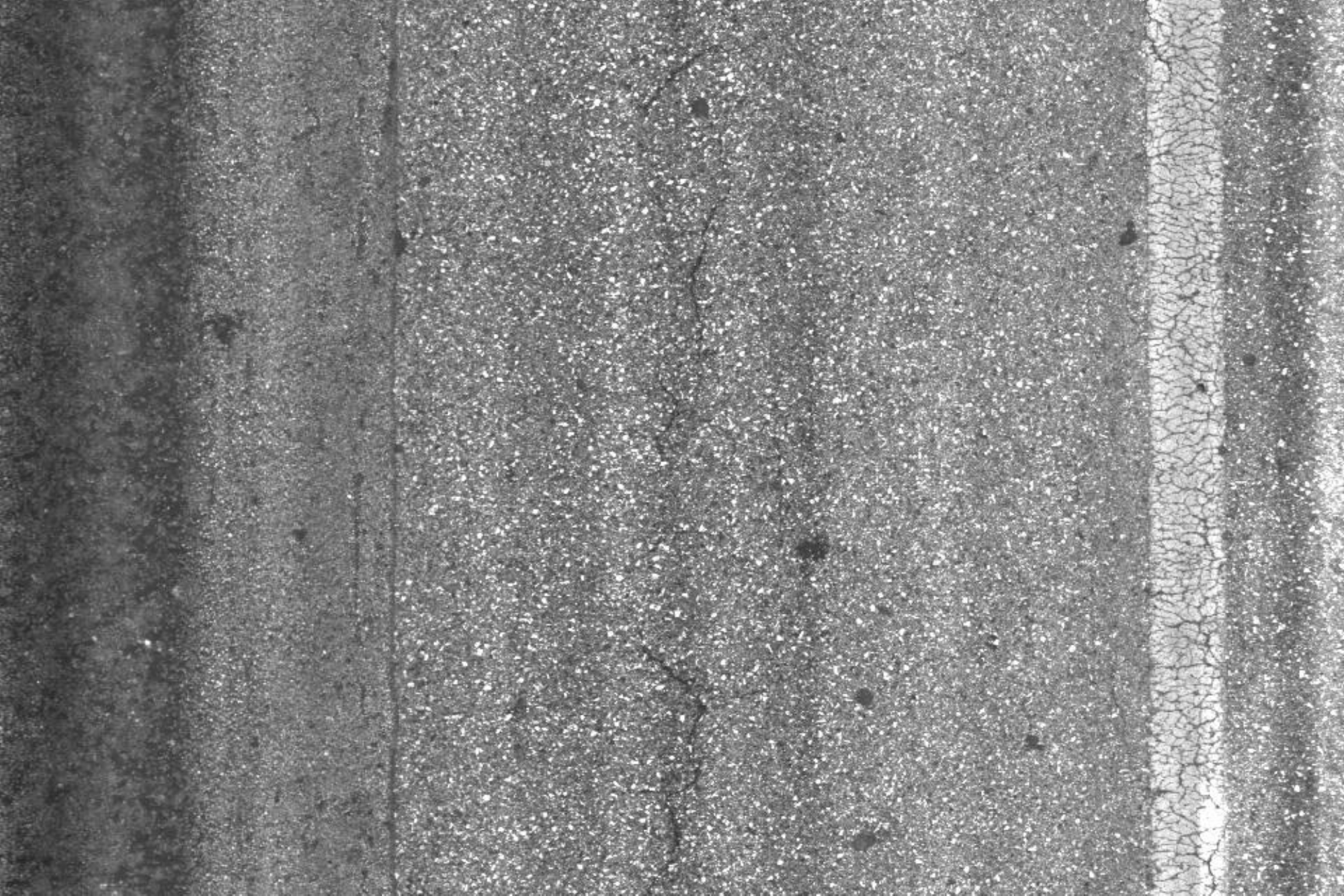}}
\subfloat{
\includegraphics[width = 0.16\textwidth]{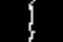}}
\subfloat{
\includegraphics[width = 0.16\textwidth]{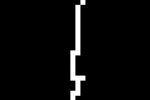}}
\subfloat{
\includegraphics[width = 0.16\textwidth]{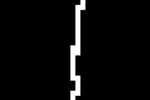}}
\subfloat{
\includegraphics[width = 0.16\textwidth]{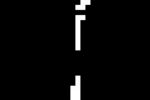}}
\subfloat{
\includegraphics[width = 0.16\textwidth]{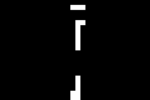}}

\subfloat{
\includegraphics[width = 0.16\textwidth]{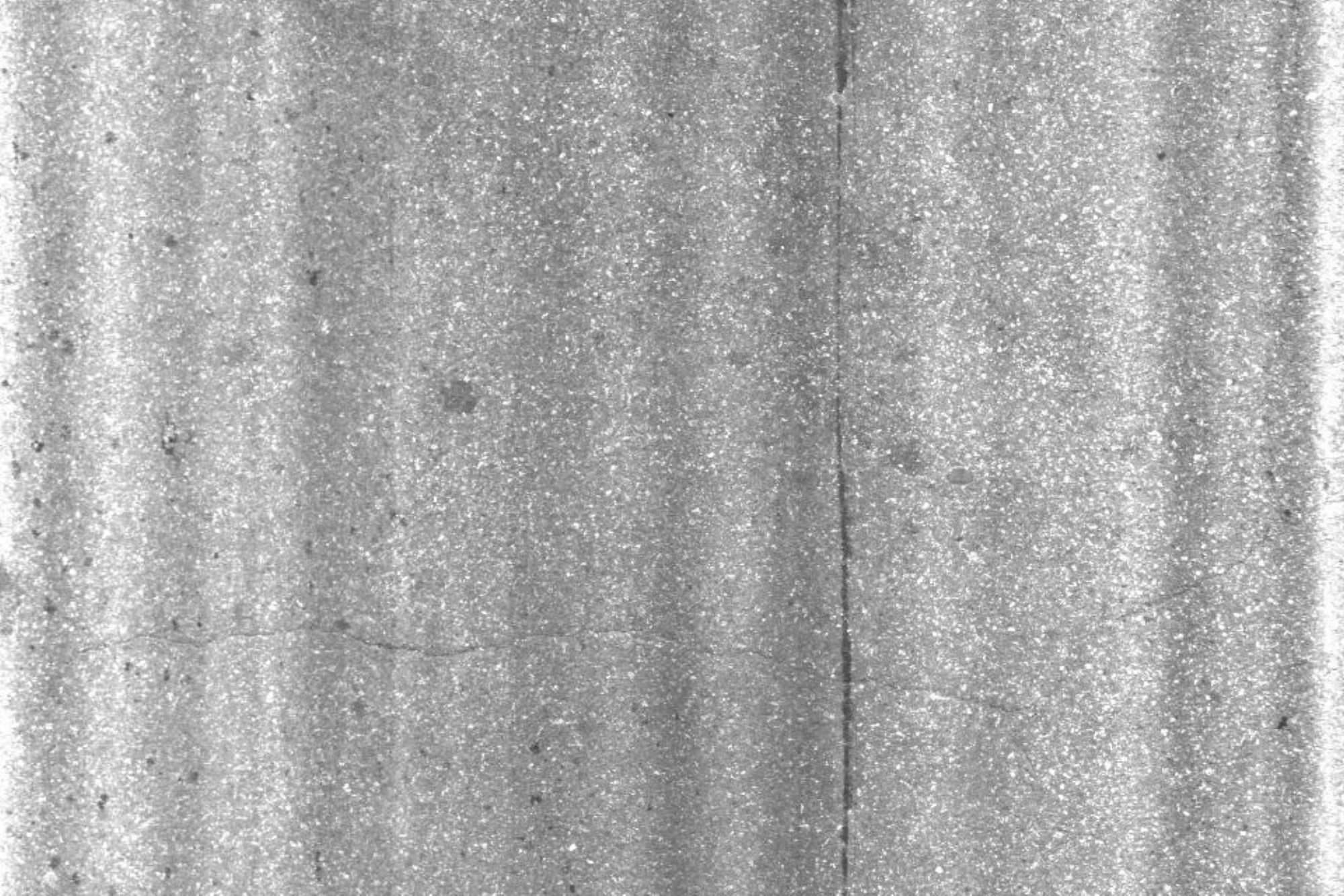}}
\subfloat{
\includegraphics[width = 0.16\textwidth]{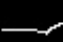}}
\subfloat{
\includegraphics[width = 0.16\textwidth]{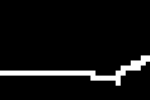}}
\subfloat{
\includegraphics[width = 0.16\textwidth]{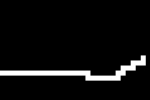}}
\subfloat{
\includegraphics[width = 0.16\textwidth]{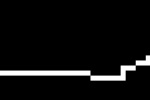}}
\subfloat{
\includegraphics[width = 0.16\textwidth]{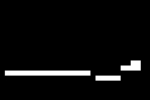}}

\subfloat{
\includegraphics[width = 0.16\textwidth]{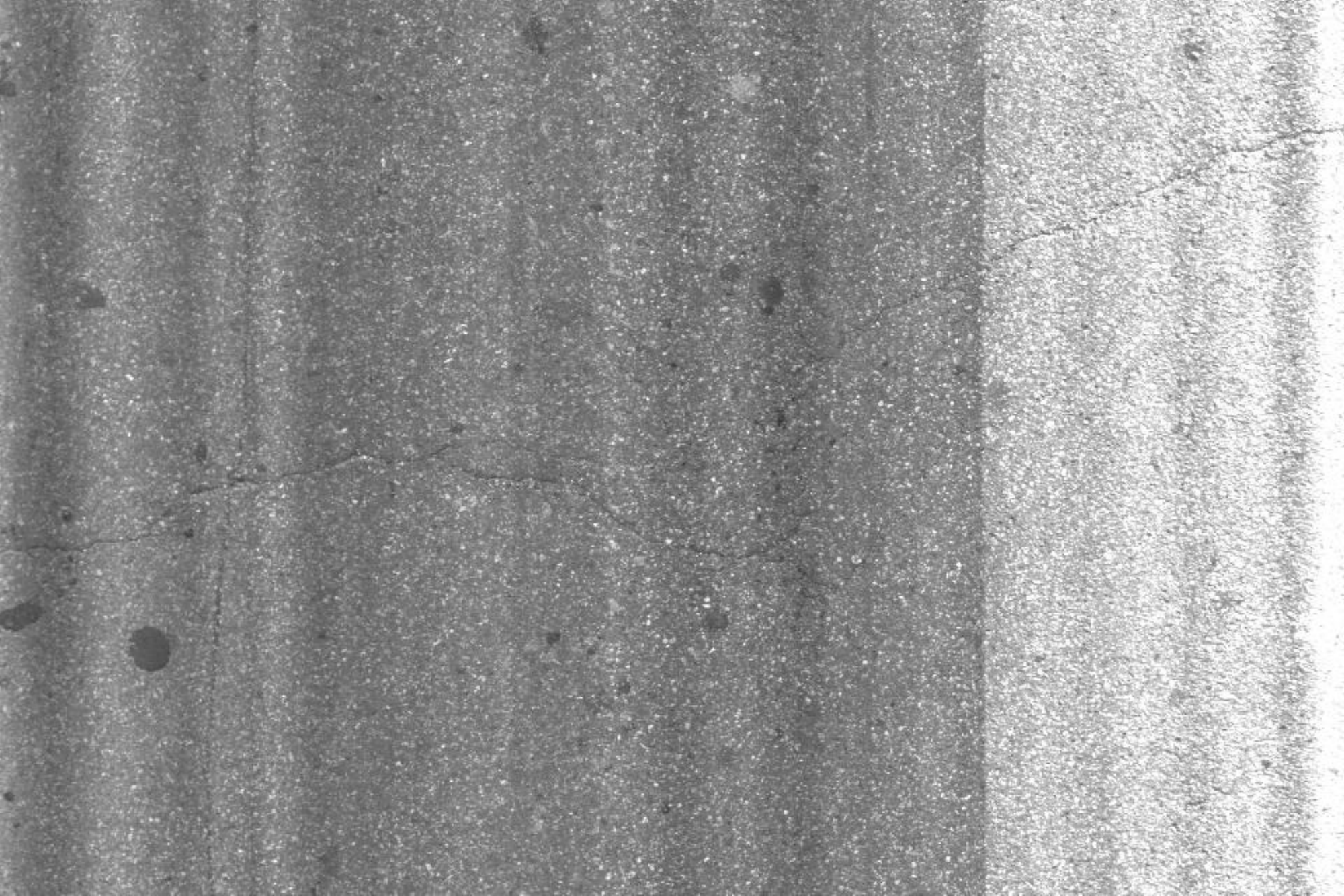}}
\subfloat{
\includegraphics[width = 0.16\textwidth]{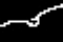}}
\subfloat{
\includegraphics[width = 0.16\textwidth]{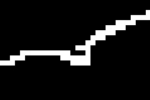}}
\subfloat{
\includegraphics[width = 0.16\textwidth]{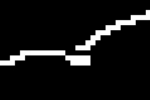}}
\subfloat{
\includegraphics[width = 0.16\textwidth]{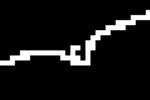}}
\subfloat{
\includegraphics[width = 0.16\textwidth]{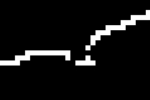}}

\setcounter {subfigure} {0}
\subfloat[Image]{
\includegraphics[width = 0.16\textwidth]{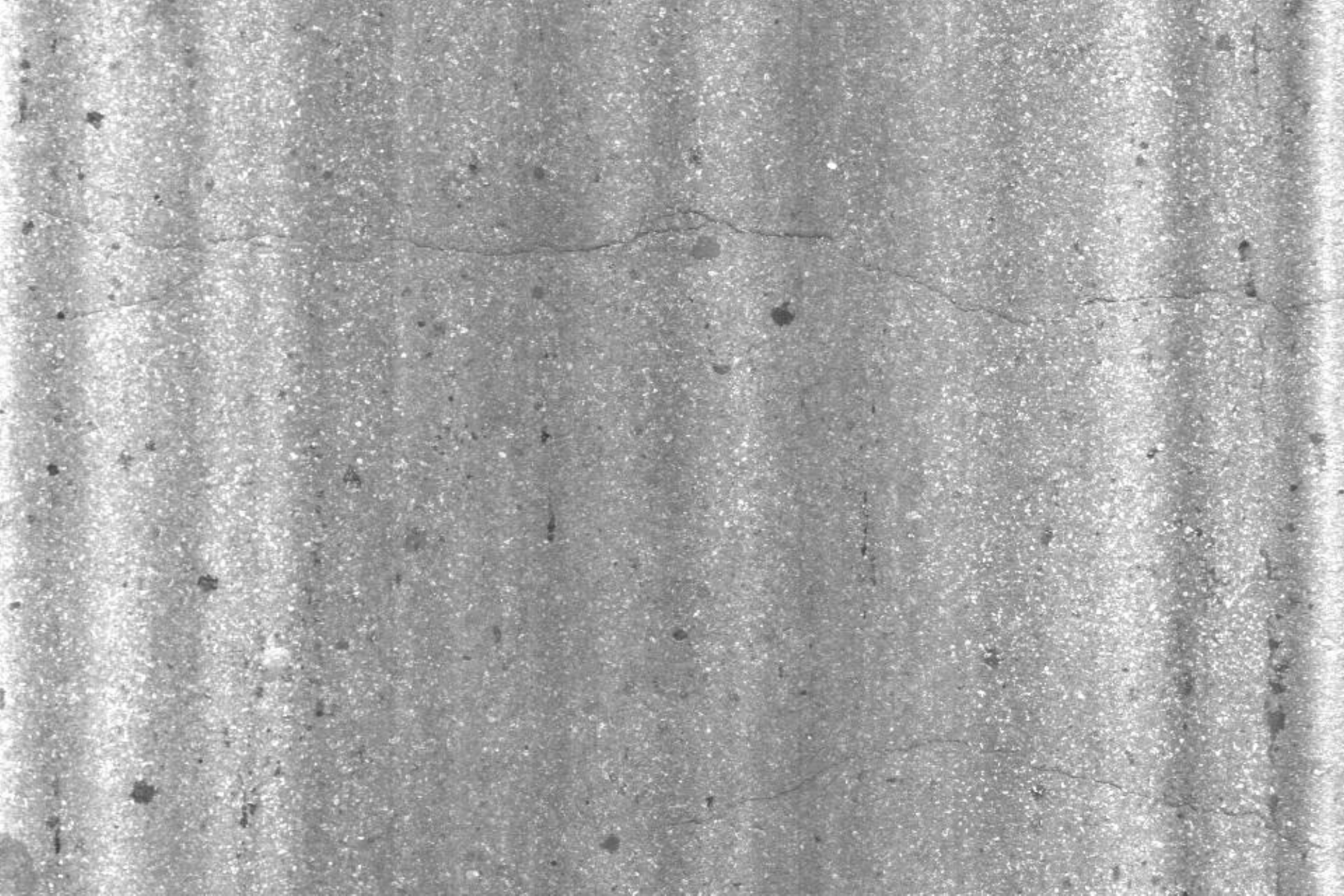}}
\subfloat[GT]{
\includegraphics[width = 0.16\textwidth]{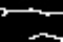}}
\subfloat[MGCrackNet]{
\includegraphics[width = 0.16\textwidth]{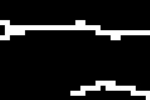}}
\subfloat[SFE]{
\includegraphics[width = 0.16\textwidth]{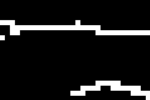}}
\subfloat[MSCNN]{
\includegraphics[width = 0.16\textwidth]{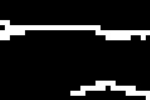}}
\subfloat[UFE]{
\includegraphics[width = 0.16\textwidth]{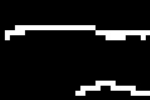}}

\caption{Visual comparisons of the STOA methods and our methods on BPC. Please zoom out for more detailed cracks.}
\label{results_visualization}
\end{center}
\end{figure*}

\begin{figure*}[!t]\scriptsize
\begin{center}
\subfloat[BPC]{\includegraphics[width=0.33\textwidth]{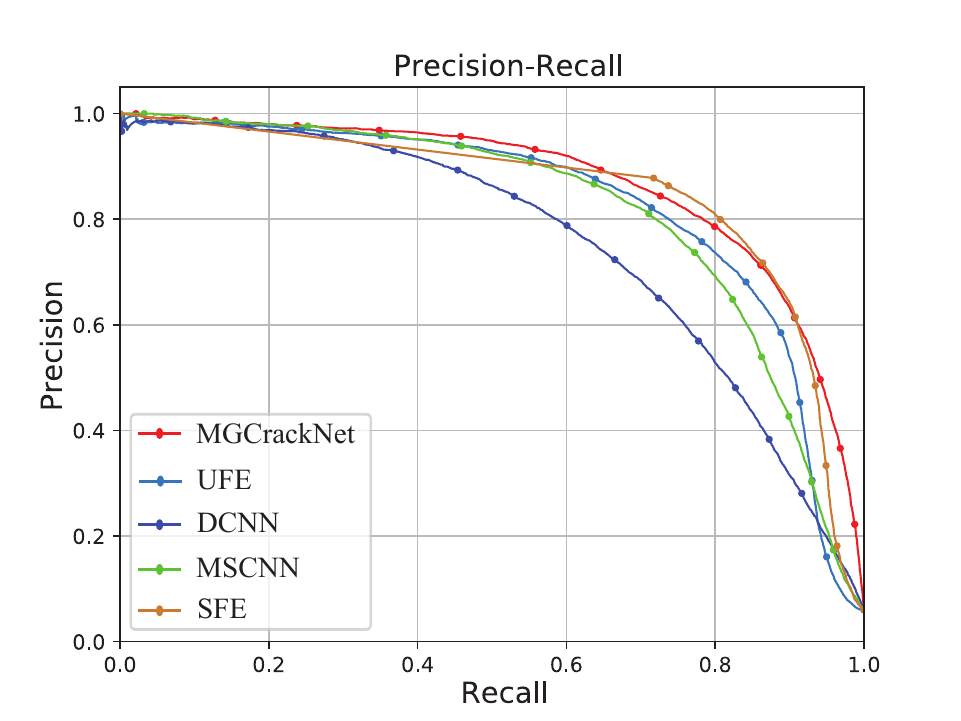}}
\subfloat[CFD]{\includegraphics[width=0.33\textwidth]{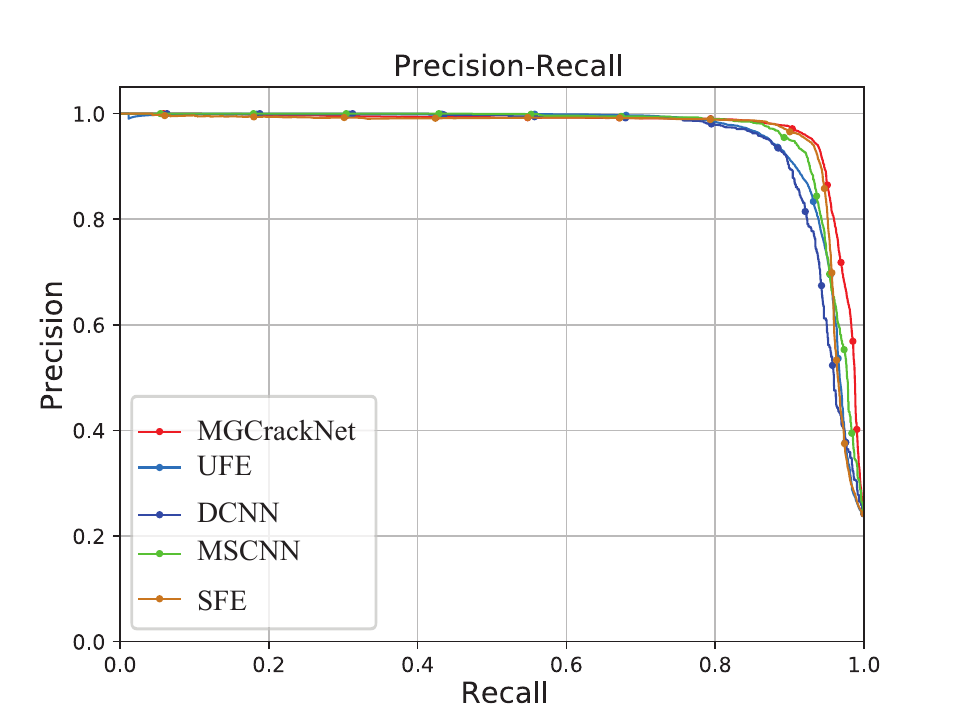}}
\subfloat[Cracktree]{\includegraphics[width=0.33\textwidth]{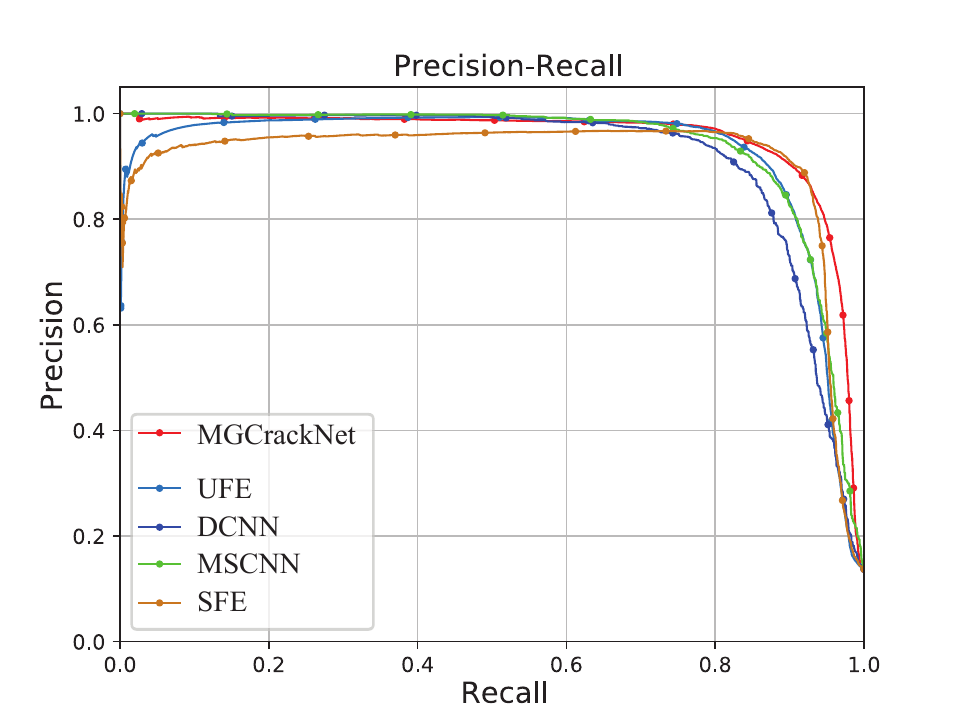}}
\caption{The Precision-Recall curves of the crack detection methods on three datasets.}
\label{img:PRcurves}
\end{center}
\end{figure*}

\subsection{Experimental Results}

\subsubsection{Results on CFD}

Tab.~\ref{experiment-result} shows that our method outperforms the other STOA methods in terms of both $AP$ and $F1$. For instance, compared with DCNN, MSCNN, UFE and SFE, our method achieves 97.29$\% AP$, which has 2.46$\%$, 0.72$\%$, 1.53$\%$ and 1.67$\%$ improvement, respectively. We notice that the CFD barely has label noise and the cracks are almost clear ones for the cement pavement. As expected, the $AP$ of MSCNN, UFE and SFE also achieve excellent performances in Tab.~\ref{experiment-result}. As a comparison, our method achieves competitive performances over two evaluation criteria from $F1$ and $AP$ metrics.

Another important observation is that UFE achieves the closest performance to our method in terms of $P$, $R$ and $F1$, as illustrated in Fig.~\ref{img:PRcurves}(b). Our method adopts the multi-stage context information flow to boost the discriminative ability of features. It indicates that multi-stage context cues is necessary for crack detection.

Either the global semantic context method (\emph{i.e.}, ST and FT) or the local one does not obtain the comparative performances in Table~\ref{experiment-result}. One possible reason is that transformer based methods perform well for a large data sets with the diversified local appearances, which help robust use attention mechanism. However, the high similarity between the background and cracks damage this assumption, especially for the BPC dataset.

\subsubsection{Results on CrackTree}

Compared to the CFD dataset, the Cracktree dataset is more complex, due to the shadows, and the blurred background.

Tab.~\ref{experiment-result} shows our method consistently achieves the best performance with an 95.24$\%$$AP$. In addition, the transformer also achieve over 80$\%$ $AP$ on Cracktree dataset, indicating that it can handle crack datasets with less noise. However, the transformer is unable to fully fit the data features due to insufficient datasets, resulting in a significant decrease in classification results compared to CNN. As illustrated in Fig.~\ref{img:PRcurves} (c), our method consistently outperforms both local features based methods and global feature based ones. It means the proposed context flow have advantage to fuse both local and global features.

\subsubsection{Results on BPC:}


Tab.~\ref{experiment-result}  shows the comparison results in terms of $F1$ and $AP$ on BPC. There are two observations in Tab.~\ref{experiment-result} as follows:
\begin{itemize}
\item Our method achieves the best performances, \emph{e.g.} 88.32$\% AP$ , significantly outperforming the other methods. Specially, the $AP$ of DCNN, MSCNN, SFE and UFE are 12.48$\%$, 8.12$\%$, 3.22$\%$, and 5.95$\%$ respectively, which are lower than that of our method. Besides, as illustrated in Fig.~\ref{img:PRcurves} that, MGCrackNet holds a curve most close to the up-right corner, achieving the best precision and recall values.
\item The recall of our method (\emph{i.e.}, 72.45$\%$) is a lower than that of UFE (\emph{i.e.}, 80.17$\%$), while the precision of our method (\emph{i.e.}, 86.90$\%$) is a higher than that of UFE (\emph{i.e.}, 76.23$\%$). One possible reason is that MGCrackNet could capture more useful context information than UFE. Besides, BPC has a lot of noisy and blurry cracks, and even some crack patches are difficult for humans to be accurately labeled. Therefore, our method has a lower recall but higher precision to guarantee a higher $AP$ than that of SFE, UFE, DCNN and MSCNN. Another possible reason is that although low-level features bring more detailed information, they also bring noisy information, especially for the stage $S_3$.
\end{itemize}

\subsection{Ablation Experiments}

\begin{table}[]
\caption{Ablation Experiments on BPC Dataset}
\centering
\renewcommand\arraystretch{1.5}
\tabcolsep=0.2cm
\begin{tabular}{cccccccc}
\hline
SPDC&CG&MIL&$P$($\%$)&$R$($\%$)&$F1$($\%$)&$AP$($\%$) \\
\hline
\checkmark & - & -  & 76.23 & \textbf{80.17} & 78.15 &82.37 \\
\checkmark  & \checkmark & -  & 85.22 & 71.03 & 77.31 & 86.62  \\
\checkmark  & \checkmark & \checkmark   & \textbf{86.90} & 72.45 & \textbf{79.02} & \textbf{88.32} \\
\hline
\end{tabular}
\label{ablation-result}
\end{table}

\begin{table}[]
\caption{THE Performance of Different Dilated Rates on BPC Dataset}
\centering
\renewcommand\arraystretch{1.5}
\tabcolsep=0.28cm
\begin{tabular}{lllll}
\hline
Dilated Rates      & $P(\%)$ & $R(\%)$ & $F1(\%$) & $AP(\%)$ \\ \hline
\{1,2,3\} & 86.27          & 71.22          & 78.02           & 86.75  \\
\{1,2,5\} & \textbf{86.90} & 72.45          & \textbf{79.02}  & \textbf{88.32}  \\
\{1,3,5\} & 85.06          & 72.22          & 78.11           & 86.96  \\
\{2,3,5\} & 80.46          & \textbf{76.86} & 78.61            & 86.37  \\ \hline
\end{tabular}
\label{ablation-rate}
\end{table}

\begin{table}[h]
\caption{THE Performance of Different Pooling Operation of MIL Module on BPC Dataset}
\centering
\renewcommand\arraystretch{1.5}
\tabcolsep=0.4cm
\begin{tabular}{ccccc}
\hline
Pooling&$P$($\%$)&$R$($\%$)&$F1$($\%$)&$AP$($\%$) \\
\hline
Max  &86.27  &72.06  &78.53  &87.67  \\
Avg  & \textbf{86.90} & \textbf{72.45} & \textbf{79.02} & \textbf{88.32}  \\
\hline
\end{tabular}
\label{ablation-mil}
\end{table}

An interesting observation in Tab.~\ref{experiment-result} is that the $AP$ of three local feature-based networks (\emph{i.e.}, DenseNet121~\cite{huang2017densely}, ResNet50~\cite{he2016deep} and VGG16~\cite{simonyan2014very}) decline the performances with the increase of network depth. It indicates that crack does not requires a larger receptive field to extract local context of cracks. Besides, the experimental results of transformer further validate the above explanation. The similarity between cracks and backgrounds make transformer~\cite{liu2021swin}~\cite{yang2021focal} inefficient for the BPC dataset. It indicates that increasing the discriminative ability of local features is more important.

\textbf{Ablation Experiments for Each Module:} As shown in Tab.~\ref{ablation-result}, the combination of each module improves the experimental results. Specifically, we first conduct experiments on the feature extraction stage of the PDC. We add the CG module to the PDC network, and the experimental results show that the gain of $AP$ increased by 4.25$\%$, demonstrating the effectiveness of that the semantic context should guides the local context. In addition, the MIL module gains a 1.7$\%$ increment in terms of $AP$, because it further optimizes the noise interference of the output results.

\textbf{Robustness of Hyper-Parameters:} To select the most appropriate hyper-parameters, we conduct a series of hyper-parameters experiments on the BPC dataset. The experimental results are shown in Tab.~\ref{ablation-rate}. The results show that the hyper-parameter $r$ = ${1, 2, 5}$ has obtained the best experimental results.

\textbf{Ablation Experiments for MIL Module:} To further validate the impact of the different pooling operations on MIL, we experimentally select the two most commonly used pooling methods: max-pooling(Max) and avg-pooling (Avg), as shown in Tab.~\ref{ablation-mil}. The avg-pooling achieves higher experimental results than that of max-pooling, gaining 0.65$\%$. The reason is that the max-pooling only outputs the value with the highest probability, which captures the information of one sub-patch. On the contrary, the avg-pooling outputs the averaged prediction, which back-propagates gradients for all sub-patches. The continuous structure of cracks is a crucial factor to determine whether the patch contains cracks or not. Therefore, avg-pooling is more suitable for crack detection.

\textbf{Visualization:} We visualize some predicted results by our model and other approaches in Fig.~\ref{results_visualization}. The images from BPC dataset contain some noises and a complex crack structure. Crack map outputted by MGCrackNet is very close to the ground truth, showing an excellent performance. As a contrary, other methods (\emph{i.e.},  SFE, UFE, and MSCNN) have discontinuous predicted results. For instance, some non-cracked blocks are incorrectly predicted. Because the connection is very important to calibrate the damage degree of roads during the post-processing stage.

\section{CONCLUSION}\label{sec:conlusion}

In this paper, we propose a MGCrackNet to handle the complex, irregular and non-discriminative cracks for bitumen pavement roads. From the perspective of leveraging context cues, multi-granularity contextual information is progressively fused by the multi-stage CG process for crack detection. In addition, we apply the MIL strategy to handle label misalignment problem. Experimental results on three crack datasets show that the proposed MGCrackNet achieves an excellent performance, demonstrating the effectiveness of our method.

Modeling context information is still an essential approach to handle crack problems. Meanwhile, we should introduce some methods to deal with both the noisy patches and labels. In the feature, how to handle noises, and how to utilize unlabeled data are interesting directions.

\bibliographystyle{IEEEtran}
\bibliography{ref}

\end{document}